\documentclass[letterpaper, 10 pt, conference]{ieeeconf}  

\IEEEoverridecommandlockouts                              

\usepackage{amsmath,amsfonts}
\usepackage{algorithmic}
\usepackage{algorithm}
\usepackage{array}
\usepackage[caption=false,font=normalsize,labelfont=sf,textfont=sf]{subfig}
\usepackage{textcomp}
\usepackage{stfloats}
\usepackage{url}
\usepackage{verbatim}
\usepackage{graphicx}
\usepackage{cite}
\usepackage{amssymb}  
\usepackage{booktabs} 
\usepackage{multirow}
\usepackage{tabularx}

\title{\LARGE \bf
Residual RL--MPC for Robust Microrobotic Cell Pushing Under Time-Varying Flow
}

\author{Yanda~Yang$^{1}$ and Sambeeta~Das$^{1}$%
\thanks{$^{1}$Y. Yang and S. Das are with the Department of Mechanical Engineering, University of Delaware, Newark, DE 19716, USA
{\tt\small \{robotyyd,samdas\}@udel.edu}.}%
}

\begin{document}
\maketitle

\maketitle
\thispagestyle{empty}
\pagestyle{empty}

\begin{abstract}
Contact-rich micromanipulation in microfluidic flow is challenging because small disturbances can break pushing contact and induce large lateral drift. We study planar cell pushing with a magnetic rolling microrobot that tracks a waypoint-sampled reference curve under time-varying Poiseuille flow in simulation. We propose a hybrid controller that augments a nominal MPC with a learned residual policy trained by SAC. The policy outputs a bounded 2D velocity correction that is contact-gated, so residual actions are applied only during robot–cell contact, preserving reliable approach behavior and stabilizing learning. All methods share the same actuation interface and speed envelope for fair comparisons. Simulation results show improved robustness and tracking accuracy over pure MPC and PID under nonstationary flow, with generalization from a clover training curve to unseen circle and square trajectories. A residual-bound sweep identifies an intermediate correction limit as the best trade-off, which we use in all benchmarks.
\end{abstract}

\section{INTRODUCTION}
\label{sec:introduction}

Microrobotic manipulation in microfluidic environments \cite{hwang2014mobile,pawashe2011two,zhang2019robotic} is a promising route for single-cell handling \cite{jager2000microrobots}, targeted transport \cite{li2018development}, and minimally invasive biomedical operations \cite{peyer2013bio}, where magnetic actuation enables wireless control at small scales \cite{yang2024quadrupole}. A central challenge, however, is that fluid disturbances and contact uncertainties can dominate dynamics at the microscale \cite{jiang2022control,yang2025microrobot}. In pushing-based tasks, even modest background-flow variation may break robot–cell contact or induce large lateral drift, leading to poor tracking and early failure.

Conventional feedback controllers such as PID and model-based designs provide useful structure and safety, but can be brittle under nonstationary disturbances and model mismatch. Model predictive control (MPC) is particularly attractive because it systematically handles constraints and finite-horizon planning \cite{pieters2016model,schwenzer2021review}. Yet, in contact-rich micromanipulation, accurate prediction is difficult due to uncertain contact transitions, hydrodynamic effects, and sensing noise, which can cause nominal plans to degrade when flow direction or magnitude changes.

A growing body of work has explored microrobotic pushing and transport in fluid using both model-based and data-driven strategies \cite{floyd2009two,pieters2015rodbot,yang2023rolling}. Early magnetic microrobot studies demonstrated autonomous manipulation in fluidic environments and emphasized that combining physics-based control with adaptive components can improve performance under fluid interactions \cite{pawashe2011two}. More recently, learning-based model estimation paired with optimization-based tracking control has been proposed to cope with hard-to-model fluid–structure interactions \cite{jia2024efficient}. In parallel, model-free geometric approaches have enabled autonomous magnetic microrobotic pushing along predefined paths using vision feedback and rule-based conditions \cite{sokolich2025autonomous}. Despite this progress, robust contact-rich curve tracking under nonstationary background flow remains challenging: small drift changes can destabilize pushing contact and trigger large cross-track excursions.

Learning-based control, especially deep reinforcement learning (RL), provides a complementary mechanism for adapting to unmodeled effects \cite{abbeel2006using,sacerdoti2024reinforcement}. Off-policy actor–critic methods such as Soft Actor-Critic (SAC) are data-efficient and effective for continuous control \cite{haarnoja2018soft}. Still, end-to-end RL can suffer from unstable exploration and unsafe behaviors during contact-critical phases \cite{cheng2019end}. A practical alternative is residual RL, where a learned correction augments a reliable nominal controller \cite{johannink2018residual}. This hybrid paradigm can combine model-based structure with data-driven adaptation while preserving a safe default behavior.

\begin{figure}[t]
    \centering
    \includegraphics[width=1.0\linewidth]{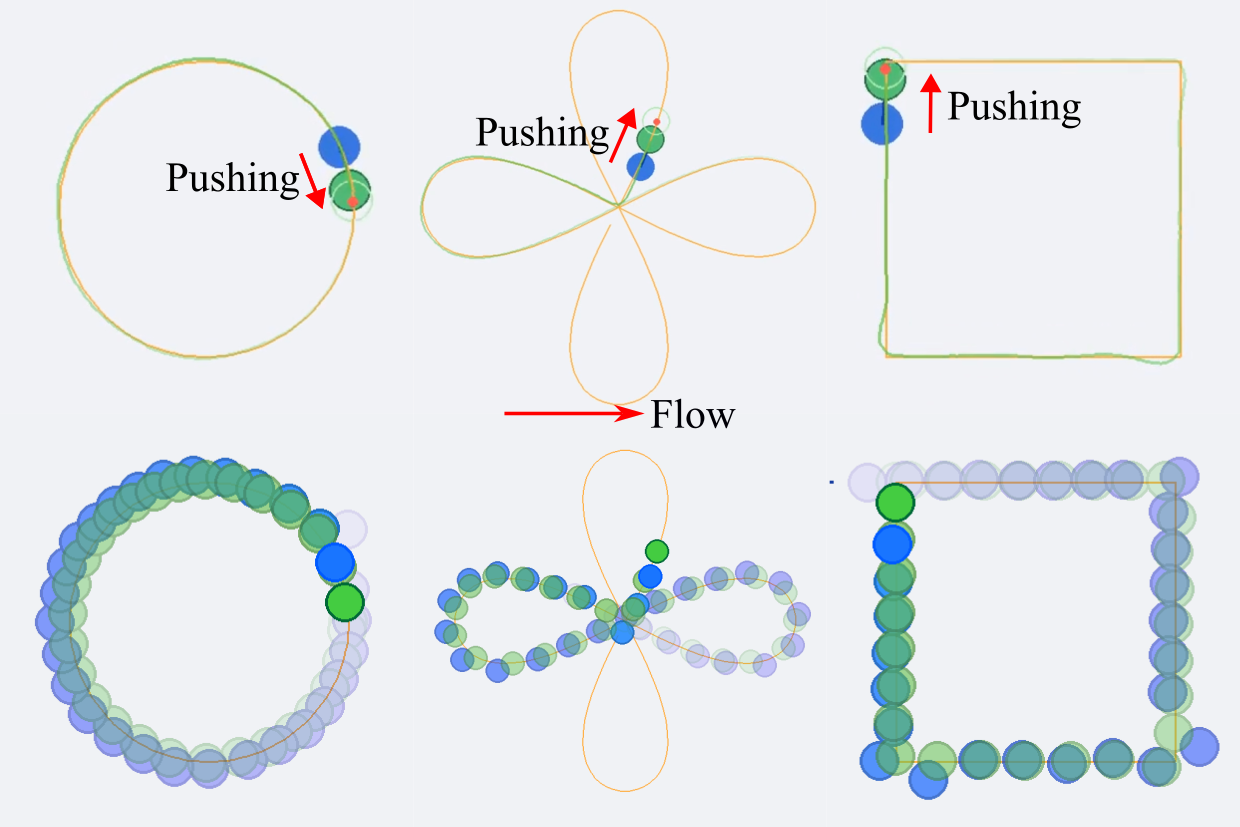}
    \caption{Task overview and evaluation curves under time-varying background flow. A magnetic rolling microrobot pushes a cell along a waypoint-sampled reference curve while compensating for drift induced by Poiseuille flow. Snapshots illustrate contact-rich pushing and typical tracking behavior under disturbance.}
    \label{fig:schematic}
\end{figure}

Motivated by these observations, we study waypoint-based cell pushing by a magnetic rolling microrobot under time-varying Poiseuille flow and propose a contact-gated residual RL–MPC controller (ResRL-MPC). The key idea is to apply a bounded learned residual velocity only during confirmed robot–cell contact, while preserving nominal MPC behavior during approach. This decomposition stabilizes learning, maintains interpretable control structure, and improves robustness to simulated time-varying disturbances (Fig.~\ref{fig:schematic}).

The implementation builds on our MicroPush simulation backend~\cite{yang2026micropush}, which provides a controlled testbed for comparing contact, actuation, and controller behavior under identical conditions. Building on this simulation foundation, the main contributions of this work are:
\begin{itemize}
    \item A contact-gated residual control architecture that augments an MPC backend with a bounded SAC policy for robust cell pushing under simulated nonstationary flow.
    \item A unified actuation interface and shared speed envelope across ResRL-MPC, pure MPC, and PID baselines for fair comparisons.
    \item A systematic simulation evaluation on seen and unseen curve geometries (Clover, Circle, Square), including a residual-bound sweep that reveals an authority–stability trade-off and motivates a practical residual limit.
\end{itemize}

\section{PROBLEM SETUP}
\label{sec:problem_setup}

We consider a contact-rich micromanipulation task in which a magnetic rolling microrobot pushes a single cell to track a prescribed planar reference curve under time-varying background microfluidic flow. Each episode consists of (i) placing the cell on the reference curve and spawning the robot behind the cell along the local curve tangent, and (ii) executing closed-loop control to advance along the curve while staying close to it.

\subsection{State and Observations}
\label{sec:state_observation}
Let the robot pose be $(\mathbf{p}_r(t),\theta_r(t))$, where $\mathbf{p}_r(t)\in\mathbb{R}^2$ and $\theta_r(t)\in\mathbb{R}$, and let the cell position be $\mathbf{p}_c(t)\in\mathbb{R}^2$. The environment includes a time-varying background flow field $\mathbf{u}_f(\mathbf{x},t)\in\mathbb{R}^2$ that induces drift disturbances during pushing.

In practice, the controller and learning policy receive a compact observation that summarizes local geometry and motion, including relative position vectors (robot-to-cell and cell-to-waypoint), robot and cell velocities, and a binary contact indicator $\mathbb{I}_{\mathrm{ct}}(t)\in\{0,1\}$ estimated either from the controller's internal contact state or from robot--cell proximity. Additional observation components and normalization details are provided in Sec.~\ref{sec:learning_setup}.

\subsection{Reference Curve and Waypoint Progression}
\label{sec:trajectory_waypoints}
The reference curve is represented by an ordered set of waypoints $\mathcal{W}=\{\mathbf{w}_i\}_{i=0}^{N-1}$ sampled from a closed curve $\gamma$. At step $k$, the active waypoint index is $i_k$ and the instantaneous goal is $\mathbf{g}_k \triangleq \mathbf{w}_{i_k}$. A waypoint is deemed reached when the cell center enters a reach region of radius $r_{\mathrm{wp}}$:
\begin{equation}
\|\mathbf{p}_c(k)-\mathbf{w}_{i_k}\|_2 \le r_{\mathrm{wp}}.
\label{eq:reach_region}
\end{equation}
Upon reaching the current waypoint, the index advances, and in some cases it advances multiple times within a single step when the cell moves far:
\begin{equation}
i_{k+1} \leftarrow (i_k + \Delta i_k)\bmod N,\qquad \Delta i_k \in \mathbb{Z}_{\ge 0}.
\label{eq:wp_advance}
\end{equation}
We track the cumulative number of advanced waypoints $A_k=\sum_{j=0}^{k}\Delta i_j$ and declare success when $A_k \ge A_{\mathrm{target}}$, with $A_{\mathrm{target}}=N$ in our simulation.

\subsection{Cross-Track Error}
\label{sec:cte}
\begin{figure}[t]
    \centering
    \includegraphics[width=0.8\linewidth]{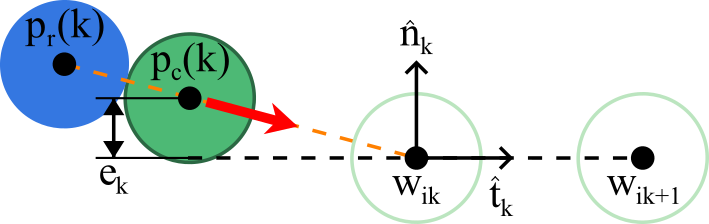}
    \caption{Local cross-track error (CTE) definition at step $k$. A local unit tangent $\hat{\mathbf{t}}_k$ is computed from the segment $\mathbf{w}_{i_k}\!\rightarrow\!\mathbf{w}_{(i_k+1)\bmod N}$, and the corresponding unit normal is $\hat{\mathbf{n}}_k$. The signed CTE is $e_k$, whose magnitude $|e_k|$ is used as the tracking error and for large-error termination.}
    \label{fig:error}
\end{figure}

To measure tracking accuracy, we define a local cross-track error (CTE) at step $k$ using the current waypoint $\mathbf{w}_{i_k}$ and the local unit tangent of the curve. Let
\begin{equation}
\hat{\mathbf{t}}_k \triangleq 
\frac{\mathbf{w}_{(i_k+1)\bmod N}-\mathbf{w}_{i_k}}
{\|\mathbf{w}_{(i_k+1)\bmod N}-\mathbf{w}_{i_k}\|_2},
\qquad
\hat{\mathbf{n}}_k \triangleq 
\begin{bmatrix}
-\hat{t}_{k,y} \\ \hat{t}_{k,x}
\end{bmatrix}.
\label{eq:tangent_normal}
\end{equation}
The signed CTE is then
\begin{equation}
e_k \triangleq \big(\mathbf{p}_c(k)-\mathbf{w}_{i_k}\big)^\top \hat{\mathbf{n}}_k,
\label{eq:cte_signed}
\end{equation}
and we use $|e_k|$ as the absolute tracking error. As illustrated in Fig.~\ref{fig:error}, this CTE is computed with respect to the local waypoint segment.

\subsection{Success and Failure Criteria}
\label{sec:success_failure}
An episode terminates successfully when the cell completes the target number of waypoint advancements:
\begin{equation}
\texttt{success} \iff A_k \ge A_{\mathrm{target}}.
\label{eq:success}
\end{equation}
We additionally enforce two termination conditions for bounded evaluation: (i) a large-error failure if the CTE exceeds a threshold $e_{\max}$,
\begin{equation}
\texttt{failure} \iff |e_k| > e_{\max},
\label{eq:failure_cte}
\end{equation}
and (ii) a truncation if the episode length reaches a maximum step budget $K_{\max}$.

\subsection{Flow Disturbance Model}
\label{sec:flow_model}
The background flow varies over time through a smooth random process that evolves a scalar centerline speed $u(t)$ within bounds $[u_{\min},u_{\max}]$. A target speed is perturbed by Gaussian noise and the current speed relaxes toward this target with time constant $\tau$, yielding a temporally correlated disturbance that induces nonstationary drift during pushing. Detailed implementation and motivation for this disturbance model are provided in Sec.~\ref{sec:learning_setup}.

\begin{table}[t]
\centering
\caption{Problem setup parameters for the trajectory-tracking task.}
\label{tab:problem_setup_params}
\small
\setlength{\tabcolsep}{4pt}
\begin{tabularx}{\columnwidth}{@{}X c c@{}}
\toprule
\textbf{Parameter} & \textbf{Symbol} & \textbf{Value} \\
\midrule
Number of waypoints on curve & $N$ & 140 \\
Waypoint reach radius & $r_{\mathrm{wp}}$ & 2 px \\
Max episode steps & $K_{\max}$ & 3000 \\
Failure threshold (CTE) & $e_{\max}$ & 10 px \\
Shared speed envelope & $v_{\max}$ & 50 px/s \\
\midrule
Flow speed bounds (centerline) & $[u_{\min},u_{\max}]$ & $[-10.8,\,10.8]$ px/s \\
Flow relaxation time constant & $\tau$ & 1.2 \\
Flow noise standard deviation & $\sigma$ & 2.0 \\
\bottomrule
\end{tabularx}
\end{table}

\section{METHOD}
\label{sec:method}

\begin{figure}[t]
    \centering
    \includegraphics[width=0.8\linewidth]{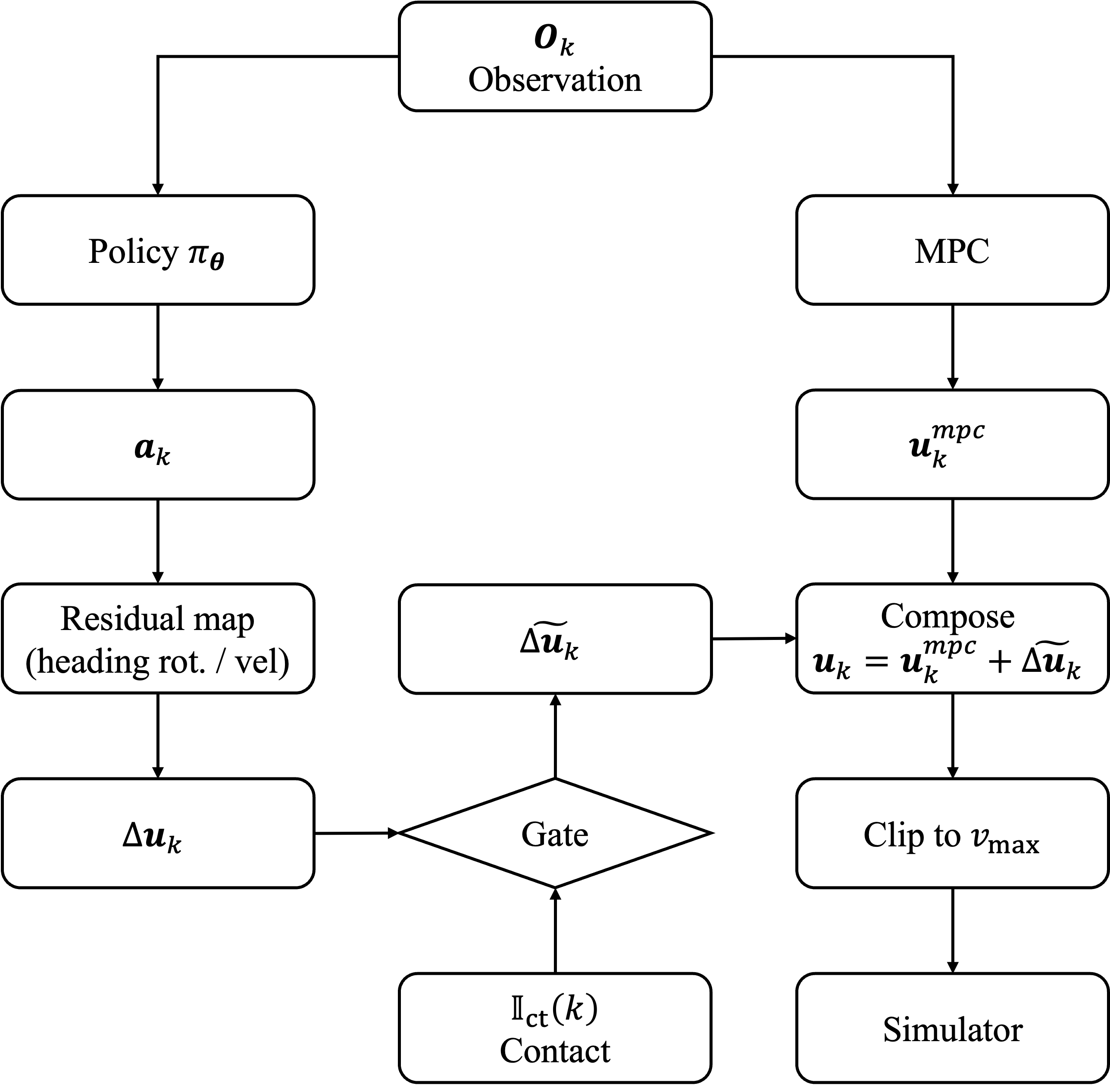}
    \caption{Contact-gated residual RL on top of MPC. The nominal MPC produces a planar velocity $\mathbf{u}^{\mathrm{mpc}}_k$. In parallel, the residual policy $\pi_\theta$ maps the observation $\mathbf{o}_k$ to a bounded action $\mathbf{a}_k\in[-1,1]^2$, which is scaled to a residual velocity $\Delta\mathbf{u}_k$ and gated by the contact indicator $\mathbb{I}_{\mathrm{ct}}(k)$. The final command is composed as $\mathbf{u}_k=\mathbf{u}^{\mathrm{mpc}}_k+\tilde{\Delta\mathbf{u}}_k$ and clipped to a shared speed envelope $v_{\max}$ before being applied to the simulator.}
    \label{fig:flowchart}
\end{figure}

We propose a hybrid controller that combines a nominal model-predictive controller (MPC) with a learned residual policy. The MPC provides structured, contact-aware pushing behavior, while the residual policy outputs bounded corrections to improve robustness under nonstationary flow and model mismatch. Crucially, the residual is contact-gated: it is activated only when the robot is in contact with the cell, which stabilizes learning and prevents erratic behaviors during the approach phase. Fig.~\ref{fig:flowchart} summarizes the computation pipeline.

\subsection{Nominal MPC Backend}
\label{sec:mpc_backend}

We use the MPC controller provided by the MicroPush control stack as the nominal backend. At each step $k$, the controller receives the current state estimate and the active goal waypoint $\mathbf{g}_k$ and produces a nominal planar command
\begin{equation}
\mathbf{u}^{\mathrm{mpc}}_k \in \mathbb{R}^2,
\label{eq:u_mpc}
\end{equation}
which represents the desired robot planar velocity in the workspace frame. The MPC incorporates a contact-aware reference strategy that biases the robot to maintain a favorable pushing configuration relative to the cell and the goal, improving stability during sustained contact.

\subsection{Residual Policy and Composition}
\label{sec:residual_composition}

Let $\pi_\theta(\cdot)$ denote a stochastic policy that maps the current observation $\mathbf{o}_k$ to a bounded 2D action
\begin{equation}
\mathbf{a}_k = \pi_\theta(\mathbf{o}_k), \qquad \mathbf{a}_k \in [-1,1]^2.
\label{eq:policy_action}
\end{equation}

\paragraph{Residual velocity parameterization (2D)}
The policy outputs a bounded action $\mathbf{a}_k$, which is mapped to a residual planar velocity correction with a shared magnitude limit:
\begin{equation}
\Delta \mathbf{u}_k = \alpha\, v_{\max}\,\mathbf{a}_k,
\label{eq:vel_residual}
\end{equation}
where $\alpha\in(0,1)$ is a scalar fraction that caps the residual magnitude relative to the common speed envelope $v_{\max}$.

\paragraph{Contact-gated composition}
Let $\mathbb{I}_{\mathrm{ct}}(k)\in\{0,1\}$ be a contact indicator. We gate the residual such that it is applied only when contact is present:
\begin{equation}
\tilde{\Delta \mathbf{u}}_k
=
\mathbb{I}_{\mathrm{ct}}(k)\,\Delta \mathbf{u}_k,
\label{eq:contact_gate}
\end{equation}
and compose the commanded planar velocity as
\begin{equation}
\mathbf{u}_k
=
\mathbf{u}^{\mathrm{mpc}}_k + \tilde{\Delta \mathbf{u}}_k.
\label{eq:u_compose}
\end{equation}
Finally, we clip $\mathbf{u}_k$ to the shared speed envelope before applying it to the simulator.

\subsection{Shared Actuation Interface and Speed Envelope}
\label{sec:shared_actuation}

All methods (pure MPC, pure PID, and our hybrid method) output a planar command $\mathbf{u}_k\in\mathbb{R}^2$. To ensure fairness and prevent any method from gaining an advantage by applying larger actuation, we enforce a shared speed envelope by clipping the command norm:
\begin{equation}
\mathbf{u}_k \leftarrow 
\mathbf{u}_k \cdot \min\!\left(1,\frac{v_{\max}}{\|\mathbf{u}_k\|_2+\epsilon}\right),
\label{eq:speed_clip}
\end{equation}
where $v_{\max}$ matches the bound used by the nominal controller and $\epsilon$ is a small constant for numerical stability. After saturation, the bounded $\mathbf{u}_k$ is applied to the simulator through the same actuation interface for all methods, so performance differences reflect decision quality rather than stronger actuation.

\subsection{Stabilizing Effect of Contact Gating}
\label{sec:gating}

The approach phase, in which the robot moves toward the cell, is sensitive to observation noise and flow-induced drift. Allowing the learned policy to act during approach can destabilize contact acquisition and increase failures by steering the robot into unfavorable configurations. By contrast, the nominal MPC is designed to establish and maintain pushing contact. Contact gating restricts learning to the contact-rich regime, enabling the policy to focus on correcting systematic errors, such as lateral drift under flow, while preserving the MPC’s reliable approach behavior. This separation improves training stability and yields robust closed-loop tracking under time-varying disturbances.

\section{LEARNING SETUP}
\label{sec:learning_setup}

This section describes learning-facing design choices: the observation provided to the residual policy, the time-varying background flow disturbance used during training and evaluation, and the reward shaping used for stable SAC training. 

\subsection{Observation Design}
\label{sec:observation_design}

The residual policy is trained to correct nominal MPC behavior under nonstationary disturbances. To make the learning problem well-conditioned, we structure the observation into three interpretable blocks: \emph{geometry}, \emph{motion}, and \emph{control context}. Geometry encodes task-relevant relative positions, motion provides local dynamics cues, and control context exposes the nominal command and tracking signals so the policy can learn when and how to correct.

\begin{table}[t]
\centering
\caption{Observation blocks provided to the residual policy.}
\label{tab:obs_blocks}
\small
\setlength{\tabcolsep}{4pt}
\begin{tabularx}{\columnwidth}{@{}l X@{}}
\toprule
\textbf{Block} & \textbf{Contents} \\
\midrule
Geometry &
Relative vectors: robot-to-cell $(\mathbf{p}_c-\mathbf{p}_r)$ and cell-to-goal $(\mathbf{g}_k-\mathbf{p}_c)$ \\
Motion &
Robot/cell planar velocities $(\mathbf{v}_r,\mathbf{v}_c)$ and robot heading $(\cos\theta_r,\sin\theta_r)$ \\
Control context &
Nominal MPC command $\mathbf{u}^{\mathrm{mpc}}_k$, contact flag $\mathbb{I}_{\mathrm{ct}}(k)$, signed cross-track error $e_k$, and local curve tangent $\hat{\mathbf{t}}_k$ \\
\bottomrule
\end{tabularx}
\end{table}

Formally, at step $k$ we define
\begin{align}
\mathbf{r}^{rc}_k &\triangleq \mathbf{p}_c(k)-\mathbf{p}_r(k)\in\mathbb{R}^2, \label{eq:obs_geom1}\\
\mathbf{r}^{cg}_k &\triangleq \mathbf{g}_k-\mathbf{p}_c(k)\in\mathbb{R}^2, \label{eq:obs_geom2}\\
\mathbf{m}_k &\triangleq \big[\mathbf{v}_r(k)^\top,\ \mathbf{v}_c(k)^\top,\ \cos\theta_r(k),\ \sin\theta_r(k)\big]^\top\in\mathbb{R}^6, \label{eq:obs_motion}
\end{align}
where $\mathbf{g}_k$ is the active waypoint. The control-context signals include the nominal MPC command $\mathbf{u}^{\mathrm{mpc}}_k\in\mathbb{R}^2$, the binary contact indicator $\mathbb{I}_{\mathrm{ct}}(k)\in\{0,1\}$, the signed cross-track error $e_k$, and the local unit tangent $\hat{\mathbf{t}}_k\in\mathbb{R}^2$:
\begin{equation}
\mathbf{c}_k \triangleq \big[(\mathbf{u}^{\mathrm{mpc}}_k)^\top,\ \mathbb{I}_{\mathrm{ct}}(k),\ (\hat{\mathbf{t}}_k)^\top,\ e_k\big]^\top \in \mathbb{R}^6.
\label{eq:obs_context}
\end{equation}
The final observation is the concatenation
\begin{equation}
\mathbf{o}_k \triangleq \big[(\mathbf{r}^{rc}_k)^\top,\ (\mathbf{r}^{cg}_k)^\top,\ \mathbf{m}_k^\top,\ \mathbf{c}_k^\top\big]^\top \in \mathbb{R}^{16}.
\label{eq:obs_concat}
\end{equation}

\paragraph{Normalization}
We first apply task-scale normalization inside the environment: position-like terms are divided by the workspace extent, velocity-like terms are divided by $v_{\max}$, and unit-vector terms are clipped to $[-1,1]$. During SAC training, we additionally apply running observation normalization to stabilize updates under varying flow conditions.

\subsection{Time-Varying Flow Disturbance}
\label{sec:flow_disturbance}

To model background microfluidic drift, the simulator applies a laminar Poiseuille profile aligned with a fixed channel axis $\hat{\mathbf{d}}$ and varying only along the cross-channel coordinate $y$:
\begin{equation}
\mathbf{u}_f(\mathbf{x},k) = u_k \left(1 - \left(\frac{y}{R}\right)^2\right)\hat{\mathbf{d}},
\label{eq:poiseuille}
\end{equation}
where $R$ is the channel half-width and $u_k$ is the centerline speed at step $k$. In practice, we modulate only the centerline speed $u_k$, while the spatial profile is handled internally by the simulator.

\begin{figure}[t]
    \centering
    \includegraphics[width=0.6\linewidth]{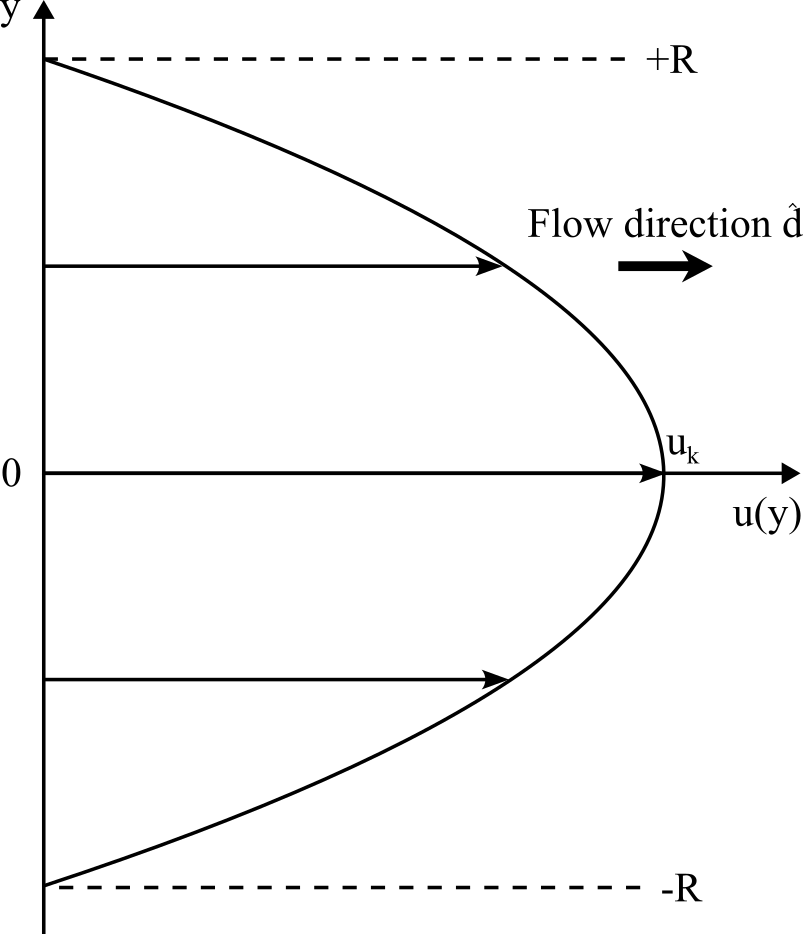}
    \caption{Poiseuille flow model used as a drift disturbance. The flow is aligned with a fixed channel axis $\hat{\mathbf{d}}$ and has a parabolic speed profile across the channel width $2R$. We randomize only the centerline speed $u_k$ over time.}
    \label{fig:poiseuille_profile}
\end{figure}

To induce nonstationary disturbances, we evolve the centerline speed using a temporally correlated random process. Let $\tilde{u}_k$ denote a clipped noisy target speed and $u_k$ the realized centerline speed. With $\Delta t = 1/f_{\mathrm{sim}}$ and $f_{\mathrm{sim}}=60$\,Hz, we update
\begin{align}
\tilde{u}_{k+1} &= \mathrm{clip}\!\left(\tilde{u}_k + \sigma \sqrt{\Delta t}\,\epsilon_k,\ u_{\min},\ u_{\max}\right),
\label{eq:flow_target}\\
&\hfill \epsilon_k \sim \mathcal{N}(0,1), \nonumber\\
u_{k+1} &= \mathrm{clip}\!\left(u_k + \alpha\left(\tilde{u}_{k+1}-u_k\right),\ u_{\min},\ u_{\max}\right),
\label{eq:flow_relax}\\
&\hfill \alpha = 1-\exp\!\left(-\frac{\Delta t}{\tau}\right). \nonumber
\end{align}
where $\tau$ is the relaxation time constant and $\sigma$ controls disturbance intensity. Training and evaluation use the same flow bounds and correlation settings (Tab.~\ref{tab:problem_setup_params}).

\subsection{Reward Shaping}
\label{sec:reward_shaping}

The residual policy is trained with a shaped reward that promotes steady advancement along the reference curve while penalizing tracking error and unnecessary corrections. At each step $k$, we use
\begin{align}
r_k &=
w_{\mathrm{prog}}\,\Delta d_k
+
w_{\mathrm{wp}}\,\Delta n^{\mathrm{wp}}_k
-
w_{\mathrm{track}}\,|e_k|
-
w_{\mathrm{time}}
\nonumber\\
&\quad
-
w_{\Delta u}\,\|\Delta \mathbf{u}_k\|_2^2
-
w_{\Delta u_s}\,\|\Delta \mathbf{u}_k-\Delta \mathbf{u}_{k-1}\|_2^2
+
r^{\mathrm{term}}_k.
\label{eq:reward_total}
\end{align}
Here, $\Delta n^{\mathrm{wp}}_k\in\mathbb{Z}_{\ge 0}$ is the number of waypoint advances at step $k$, and $e_k$ is the signed cross-track error. The local progress term is the reduction in distance to the active waypoint,
\begin{equation}
\Delta d_k \triangleq d^{\mathrm{wp}}_{k-1}-d^{\mathrm{wp}}_{k}, 
\qquad
d^{\mathrm{wp}}_{k} \triangleq \|\mathbf{p}_c(k)-\mathbf{w}_{i_k}\|_2 .
\label{eq:local_progress}
\end{equation}
We define the applied residual correction as the deviation between the executed planar command, after contact gating and any saturation, and the nominal MPC command,
\begin{equation}
\Delta \mathbf{u}_k \triangleq \mathbf{u}_k - \mathbf{u}^{\mathrm{mpc}}_k,
\label{eq:residual_u_def}
\end{equation}
which becomes zero when contact-gating disables the residual, and may be reduced when the shared speed envelope clips the composed command.

The terminal reward is
\begin{equation}
r^{\mathrm{term}}_k =
\begin{cases}
+r_{\mathrm{succ}}, & \text{if success},\\
-r_{\mathrm{fail}}, & \text{if failure (large CTE)},\\
0, & \text{if truncated at } K_{\max}.
\end{cases}
\label{eq:reward_terminal}
\end{equation}

\paragraph{Interpretation}
The progress and waypoint terms encourage consistent advancement; the tracking penalty limits lateral drift; the time penalty discourages dithering; and the residual magnitude and smoothness penalties regularize the learned correction to remain small and non-oscillatory. All coefficients are listed in Tab.~\ref{tab:reward_weights}.

\begin{table}[t]
\centering
\caption{Reward terms and coefficients used in this work.}
\label{tab:reward_weights}
\small
\setlength{\tabcolsep}{4pt}
\begin{tabularx}{\columnwidth}{@{}X l@{}}
\toprule
\textbf{Term} & \textbf{Coefficient} \\
\midrule
Local progress & $w_{\mathrm{prog}} = 5.0$ \\
Waypoint advance & $w_{\mathrm{wp}} = 3.0$ \\
Cross-track penalty & $w_{\mathrm{track}} = 0.12$ \\
Time penalty & $w_{\mathrm{time}} = 0.002$ \\
Residual magnitude penalty & $w_{\Delta u} = 0.005$ \\
Residual smoothness penalty & $w_{\Delta u_s} = 0.002$ \\
Success reward & $r_{\mathrm{succ}} = 8.0$ \\
Failure penalty & $r_{\mathrm{fail}} = 2.0$ \\
\bottomrule
\end{tabularx}
\end{table}

\section{SIMULATION EVALUATION}
\label{sec:SIMULATION EVALUATION}

We evaluate contact-gated residual RL on a contact-rich cell-pushing task with time-varying background flow. Our simulation studies address: (i) whether the residual policy improves robustness over nominal controllers under nonstationary disturbances, and (ii) whether the learned correction generalizes to unseen curve geometries when training is restricted to a single curve family.

\subsection{Training Protocol}
\label{sec:training_protocol}

\paragraph{Algorithm and implementation}
We train the residual policy using Soft Actor-Critic (SAC) with an MLP actor--critic ($[256,256]$ hidden units) for $5\times 10^5$ environment steps. Observations are normalized in two stages: (i) task-scale normalization inside the environment, and (ii) running mean/variance normalization during training using a vectorized normalizer whose statistics are frozen and reused at evaluation. Reward normalization is enabled during training for stability and disabled during evaluation.

\begin{figure}[t]
    \centering
    \includegraphics[width=0.8\linewidth]{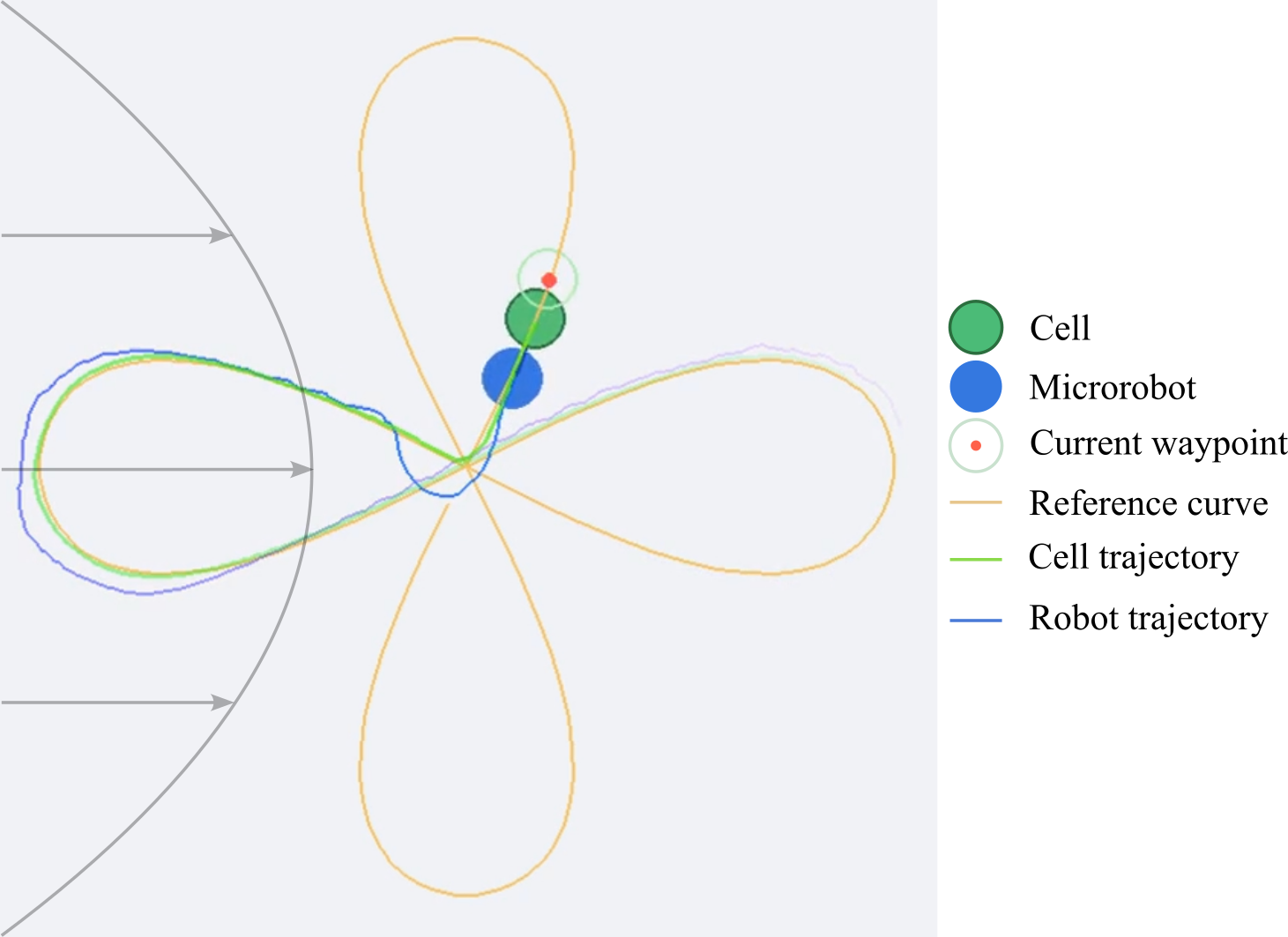}
    \caption{Training task snapshot (clover curve) under time-varying background flow. The cell (green) is pushed by the microrobot (blue) to follow the reference curve (orange). The current waypoint is shown with a marker and reach region. The cell and robot trajectories are overlaid to illustrate tracking behavior under drift.}
    \label{fig:task_snapshot}
\end{figure}

\paragraph{Training task distribution (clover curve)}
Training uses a compact clover reference curve that repeatedly changes curvature and direction within a small workspace, creating frequent contact-rich interactions under flow. The curve is implemented as a composite of two orthogonal lemniscate segments. Let a base lemniscate be
\begin{equation}
\gamma_0(t) =
\begin{bmatrix}
a\sin t\\
b\sin t\cos t
\end{bmatrix}, \qquad t\in[0,2\pi),
\label{eq:lemniscate}
\end{equation}
and let $\mathcal{R}_{90}$ denote a $90^\circ$ rotation. The full clover trajectory is generated by concatenating $\gamma_0(t)$ and $\mathcal{R}_{90}\gamma_0(t)$ with consistent orientation, then sampling $N$ waypoints uniformly in the curve parameter to form $\{\mathbf{w}_i\}_{i=0}^{N-1}$. Fig.~\ref{fig:task_snapshot} shows a snapshot of the task and visualization conventions.

During training, we randomize initial conditions and geometry to prevent overfitting: curve center jitter and curve scale parameters, randomized starting phase along the curve, and small pose jitter at reset. Background flow is always enabled and evolves according to Sec.~\ref{sec:flow_disturbance}, with centerline speed initialized uniformly in a bounded range and then updated by the temporally correlated process.

\paragraph{Hyperparameters}
Tab.~\ref{tab:sac_hparams} summarizes key SAC hyperparameters and evaluation settings used during training.

\begin{table}[t]
\centering
\caption{SAC training hyperparameters and implementation settings.}
\label{tab:sac_hparams}
\small
\setlength{\tabcolsep}{4pt}
\begin{tabularx}{\columnwidth}{@{}l X@{}}
\toprule
\textbf{Hyperparameter} & \textbf{Value} \\
\midrule
Total training steps & $5\times 10^5$ \\
Policy / critic network & MLP, $[256,256]$ \\
Learning rate & $3\times 10^{-4}$ \\
Replay buffer size & $1\times 10^5$ \\
Learning starts & $5\times 10^3$ steps \\
Batch size & 256 \\
Discount factor $\gamma$ & 0.99 \\
Target smoothing $\tau$ & 0.005 \\
Train frequency & 1 step \\
Gradient steps per update & 2 \\
Observation normalization & running mean/variance (frozen at eval) \\
Reward normalization & train: on, eval: off \\
Evaluation during training & every 5k steps, 10 episodes \\
\bottomrule
\end{tabularx}
\end{table}

\subsection{Model Selection: Residual Bound Fraction $\alpha$}
\label{sec:alpha_model_selection}

\begin{figure}[t]
    \centering
    \includegraphics[width=1.0\linewidth]{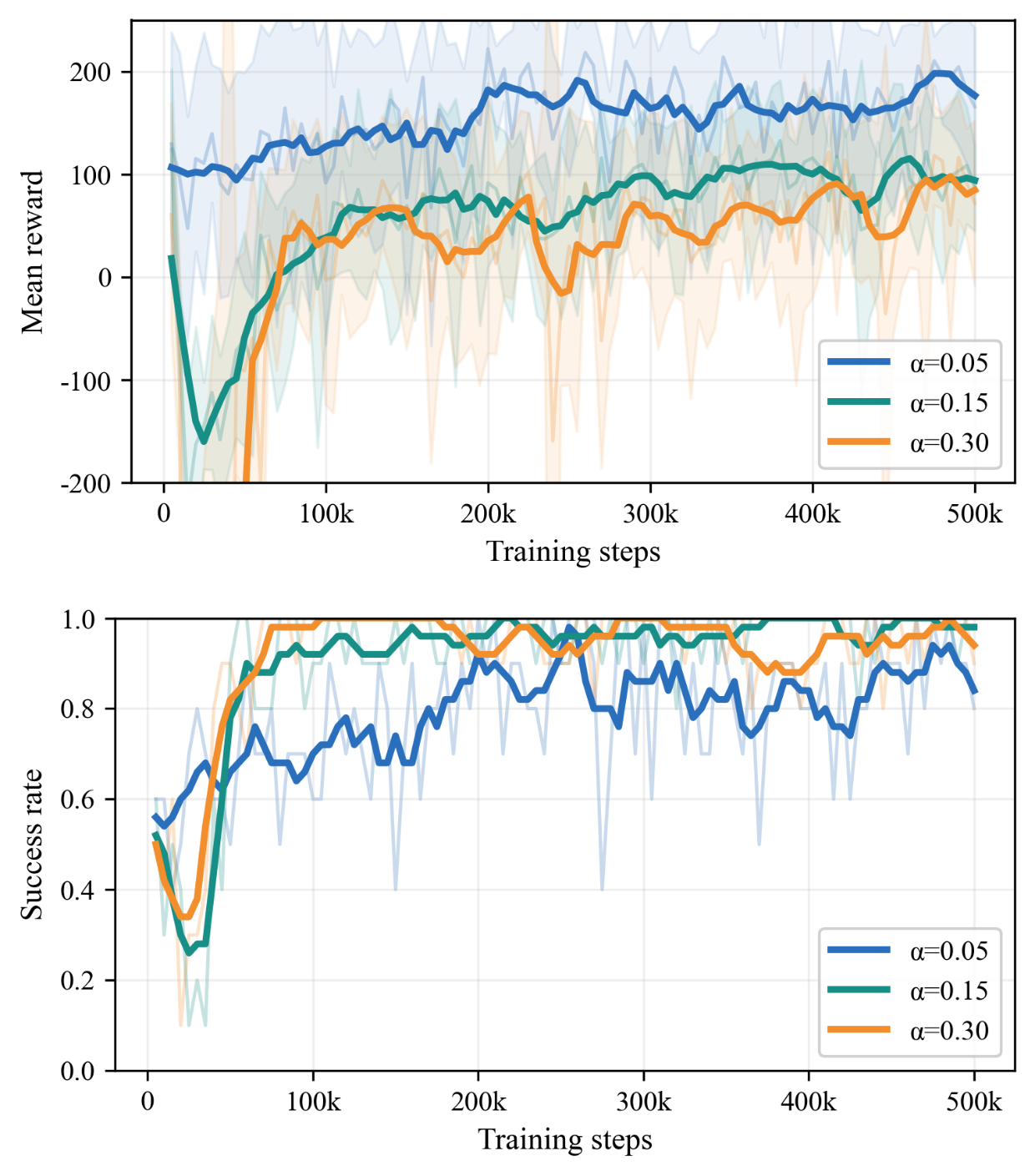}
    \caption{Residual bound sweep for model selection. We train three policies with $\alpha\in\{0.05,0.15,0.30\}$ and report evaluation mean return (top) and success rate (bottom) during training. Light traces show raw evaluation values; dark traces show a smoothed curve.}
    \label{fig:alpha_training}
\end{figure}

We select the residual bound fraction $\alpha$ in Eq.~\eqref{eq:vel_residual}, which limits the maximum residual correction as $\|\Delta \mathbf{u}\|_2 \le \alpha v_{\max}$. We train three SAC policies for $5\times 10^5$ steps each with $\alpha\in\{0.05,\,0.15,\,0.30\}$ while keeping all other settings fixed. Fig.~\ref{fig:alpha_training} reports evaluation mean return and success rate during training (raw traces with light lines and a smoothed curve for readability). Tab.~\ref{tab:alpha_ablation} summarizes final performance.

\begin{table}[t]
\centering
\caption{Residual bound sweep results. $n=20$ episodes are evaluated for each $\alpha$. Completion time and CTE are computed over successful episodes only; progress ratio is computed over all episodes.}
\label{tab:alpha_ablation}
\small
\setlength{\tabcolsep}{6pt}
\begin{tabular}{c c c c c c}
\toprule
$\alpha$ & $n$ & Success & Time$_{\text{succ}}$ (s) & CTE$_{\text{succ}}$ (px) & Progress \\
\midrule
0.05 & 20 & 0.10 & 19.87 & 0.951 & 0.299 \\
0.15 & 20 & 1.00 & 19.36 & 0.737 & 1.000 \\
0.30 & 20 & 0.85 & 21.56 & 0.698 & 0.920 \\
\bottomrule
\end{tabular}
\end{table}

A very small bound ($\alpha=0.05$) under-corrects flow-induced drift, leading to frequent failures and low progress. Increasing the bound to $\alpha=0.15$ yields the best overall performance, achieving high success with improved tracking. A larger bound ($\alpha=0.30$) remains competitive in tracking error but reduces reliability and efficiency, consistent with occasional over-correction under disturbance. Based on this sweep, we use $\alpha=0.15$ for all main comparisons against nominal baselines below.

\subsection{Evaluation Protocol}
\label{sec:evaluation_protocol}

\paragraph{Train--test split and generalization}
We train on the clover curve family (Fig.~\ref{fig:task_snapshot}) and evaluate on three curve types: Circle, Clover (seen), and Square. During evaluation, we disable geometry randomization and initialization jitter, using fixed curve parameters and deterministic resets, to ensure reproducible comparisons across methods. Flow remains time-varying during evaluation under the same disturbance regime as training, with matched bounds, correlation time, and noise level.

\paragraph{Baselines and shared actuation envelope}
We compare against two nominal controllers that share the same actuation interface and limits: (i) pure MPC, with the residual forced to zero while keeping the same control pipeline, and (ii) pure PID, evaluated under the same speed envelope. Improvements therefore reflect better decision-making under identical actuation constraints.

\paragraph{Metrics}
We report four metrics per curve over multiple random seeds:
\begin{itemize}
    \item \textbf{Success rate:} fraction of episodes that complete one lap.
    \item \textbf{Completion time:} time-to-success, computed over successful episodes only.
    \item \textbf{Tracking error:} mean absolute CTE, $\frac{1}{K}\sum_{k=1}^{K}|e_k|$.
    \item \textbf{Progress ratio:} waypoint advancement at termination normalized by the one-lap target, $\frac{A_K}{A_{\mathrm{target}}}$, computed over all episodes.
\end{itemize}

\paragraph{Implementation details}
For the learned method, we load the observation-normalization statistics saved during training and disable further updates during evaluation. All controllers are evaluated under identical environment settings, including flow bounds, $\tau$, $\sigma$, waypoint resolution, and termination thresholds.

\section{RESULTS AND DISCUSSION}
\label{sec:results}

We report quantitative comparisons across Circle, Clover (train), and Square under time-varying flow, and complement them with qualitative trajectory and failure analyses. Our goal is to evaluate whether contact-gated residual RL improves robustness and tracking accuracy while operating under the same actuation envelope as nominal controllers. ResRL-MPC denotes the hybrid method with $\alpha=0.15$ selected in Sec.~\ref{sec:alpha_model_selection}.

\subsection{Quantitative Results}
\label{sec:quant_results}

\begin{figure*}[t]
    \centering
    \includegraphics[width=0.8\linewidth]{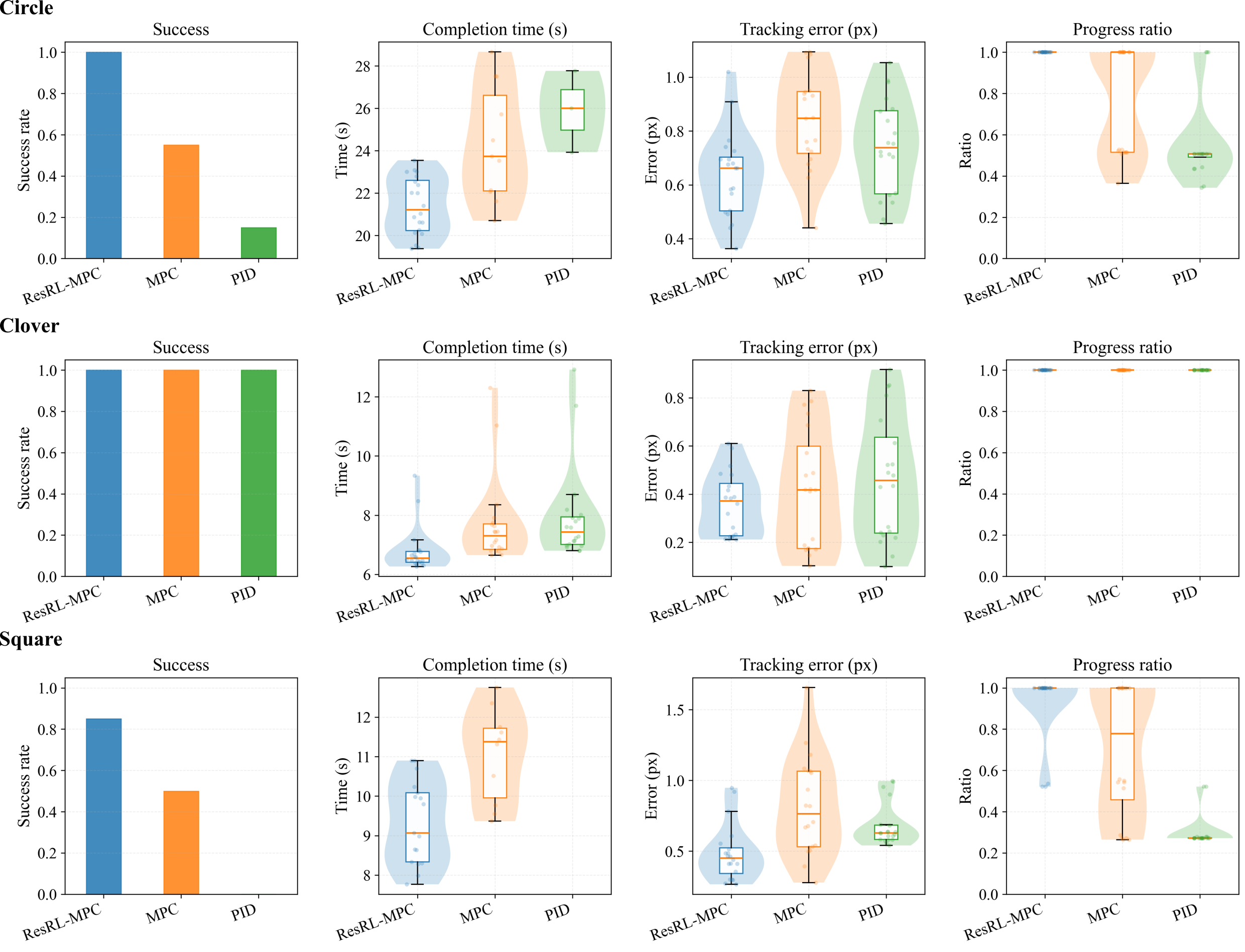}
    \caption{Quantitative results under time-varying flow for Circle (top), Clover (middle, train), and Square (bottom), aggregated over 20 seeds. We compare ResRL-MPC against pure MPC and pure PID using success rate, completion time, tracking error (CTE), and progress ratio. Completion time and tracking error are computed over successful episodes only; progress ratio is computed over all episodes.}
    \label{fig:quant_results}
\end{figure*}

Fig.~\ref{fig:quant_results} summarizes performance across 20 seeds using success rate, completion time, tracking error, and progress ratio. Overall, ResRL-MPC achieves higher success and lower tracking error than pure MPC and PID, with the largest gains on the more challenging curves. For successful trials, ResRL-MPC typically matches or improves completion time while reducing CTE, indicating more efficient disturbance rejection during sustained contact.

Progress ratio provides additional resolution beyond binary success. Pure MPC/PID often terminate early on Clover and Square, resulting in low progress ratios. In contrast, ResRL-MPC maintains higher progress even in unsuccessful trials, suggesting more consistent advancement before termination.

\subsection{Qualitative Analysis and Failure Modes}
\label{sec:qual_results}

\begin{figure*}[t]
    \centering
    \includegraphics[width=0.8\linewidth]{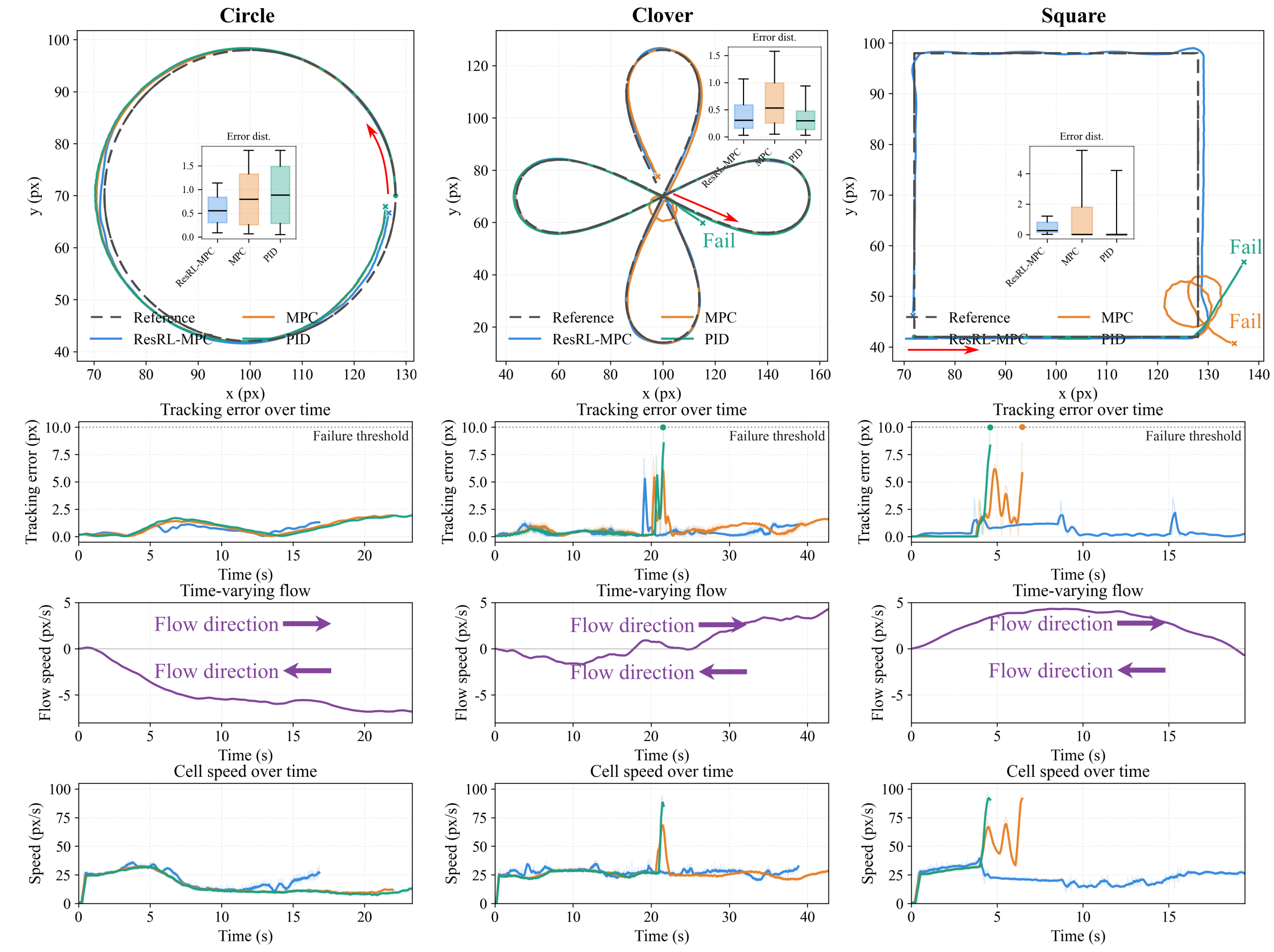}
    \caption{Qualitative rollouts under time-varying flow for Circle (left), Clover (middle, train), and Square (right). Top row: reference curve (dashed) and executed trajectories; insets show tracking error distributions. Middle rows: tracking error and realized flow speed over time. Bottom row: cell speed over time. Baseline failures correlate with sharp error spikes, whereas ResRL-MPC suppresses these spikes through a bounded contact-phase correction under the same speed envelope.}
    \label{fig:qual_results}
\end{figure*}

Fig.~\ref{fig:qual_results} shows representative rollouts, tracking error, flow speed, and cell speed for each curve; corresponding simulated rollouts are provided in the supplementary video. Baseline failures typically occur when flow changes or high-curvature segments induce sharp tracking-error spikes above the failure threshold. ResRL-MPC suppresses these spikes by applying the bounded residual correction in Eq.~\ref{eq:vel_residual} only during contact, improving lateral disturbance rejection without additional actuation authority. During approach, the residual is gated off; during sustained contact, it adjusts the commanded velocity direction relative to the nominal MPC command to maintain a favorable pushing configuration. 

\section{CONCLUSIONS}
\label{sec:conclusion}
We studied contact-rich cell pushing with a magnetic rolling microrobot under simulated time-varying microfluidic flow, formulated as waypoint-based tracking of a planar reference curve. We proposed ResRL-MPC, which augments a contact-aware MPC backend with a contact-gated residual policy that outputs bounded 2D velocity corrections. Using a shared actuation interface and speed envelope, performance gains reflect improved decision-making rather than stronger actuation.

Across Circle, Clover, and Square trajectories under the same simulated nonstationary disturbance process, ResRL-MPC improves success rate, progress ratio, and tracking accuracy relative to pure MPC and PID. A residual-bound sweep further shows that an intermediate correction limit yields the best authority–stability trade-off.

This study provides a controlled simulation-based evaluation of contact-gated residual control for microrobotic cell pushing. Future work will validate the framework on a physical magnetic actuation system in microfluidic chips with live imaging and controlled background flow.

\section{ACKNOWLEDGMENTS}
This work was supported by the National Science Foundation under grant GCR 2219101, CPS 315 2234869, and the National Health Institute under grant 1R35GM147451.  The authors acknowledge the use of ChatGPT (OpenAI) for English-language editing, code drafting, refactoring, and plotting support. All technical content, simulation design, results, and conclusions were produced and verified by the authors.






\bibliographystyle{IEEEtran}
\bibliography{references}

\end{document}